\documentclass[11pt,a4paper]{article}
\usepackage[hyperref]{acl2023}
\usepackage{times}
\usepackage{smile}
\usepackage{adjustbox}
\usepackage{latexsym}

\usepackage{microtype}

\aclfinalcopy 

\newcommand\ModelName{REVISE}

\title{Interactive Editing for Text Summarization}

\author{Yujia Xie $\quad$ Xun Wang $\quad$ Si-Qing Chen $\quad$ Wayne Xiong $\quad$ Pengcheng He  \\
  Microsoft \\
  \texttt{\{yujiaxie, xunwang, sqchen, Wayne.Xiong, Pengcheng.H\}@microsoft.com}
  }

\date{}

\begin{document}
\maketitle
\begin{abstract}
    
    Summarizing lengthy documents is a common and essential task in our daily lives. Although recent advancements in neural summarization models can assist in crafting general-purpose summaries, human writers often have specific requirements that call for a more customized approach. To address this need, we introduce \ModelName, an innovative framework designed to facilitate iterative editing and refinement of draft summaries by human writers. Within our framework, writers can effortlessly modify unsatisfactory segments at any location or length and provide optional starting phrases -- our system will generate coherent alternatives that seamlessly integrate with the existing summary.
    At its core, \ModelName~incorporates a modified fill-in-the-middle model with the encoder-decoder architecture while developing novel evaluation metrics tailored for the summarization task. 
    In essence, our framework empowers users to create high-quality, personalized summaries by effectively harnessing both human expertise and AI capabilities, ultimately transforming the summarization process into a truly collaborative and adaptive experience.
    
\end{abstract}
\section{Introduction}

Human intelligence has been significantly augmented by the rapid development of Artificial Intelligence (AI, \citet{engelbart1962augmenting, lee2022evaluating}), particularly with the emergence of large language models \citep{devlin2018bert, raffel2020exploring, brown2020language, OpenAI2023GPT4TR}. AIs have shown great potential in various practical settings, helping users brainstorm ideas (e.g., \href{https://www.jasper.ai/}{Jasper}, \href{https://copy.ai/}{Copy.ai}), paraphrase sentences (e.g.,\href{https://www.wordtune.com/}{Wordtune}, \href{https://quillbot.com/}{QuillBot}), reformulate queries \citep{nogueira2017task}, autocomplete sentences \citep{chen2019gmail}, and write code (e.g., \href{https://github.com/features/copilot}{Copilot}, \href{https://www.tabnine.com/}{TabNine}). One specific area where AI can revolutionize our daily life is in the realm of document summarization, which is the focus of this paper.

\begin{figure}[t]
  \centering  \includegraphics[width=\linewidth]{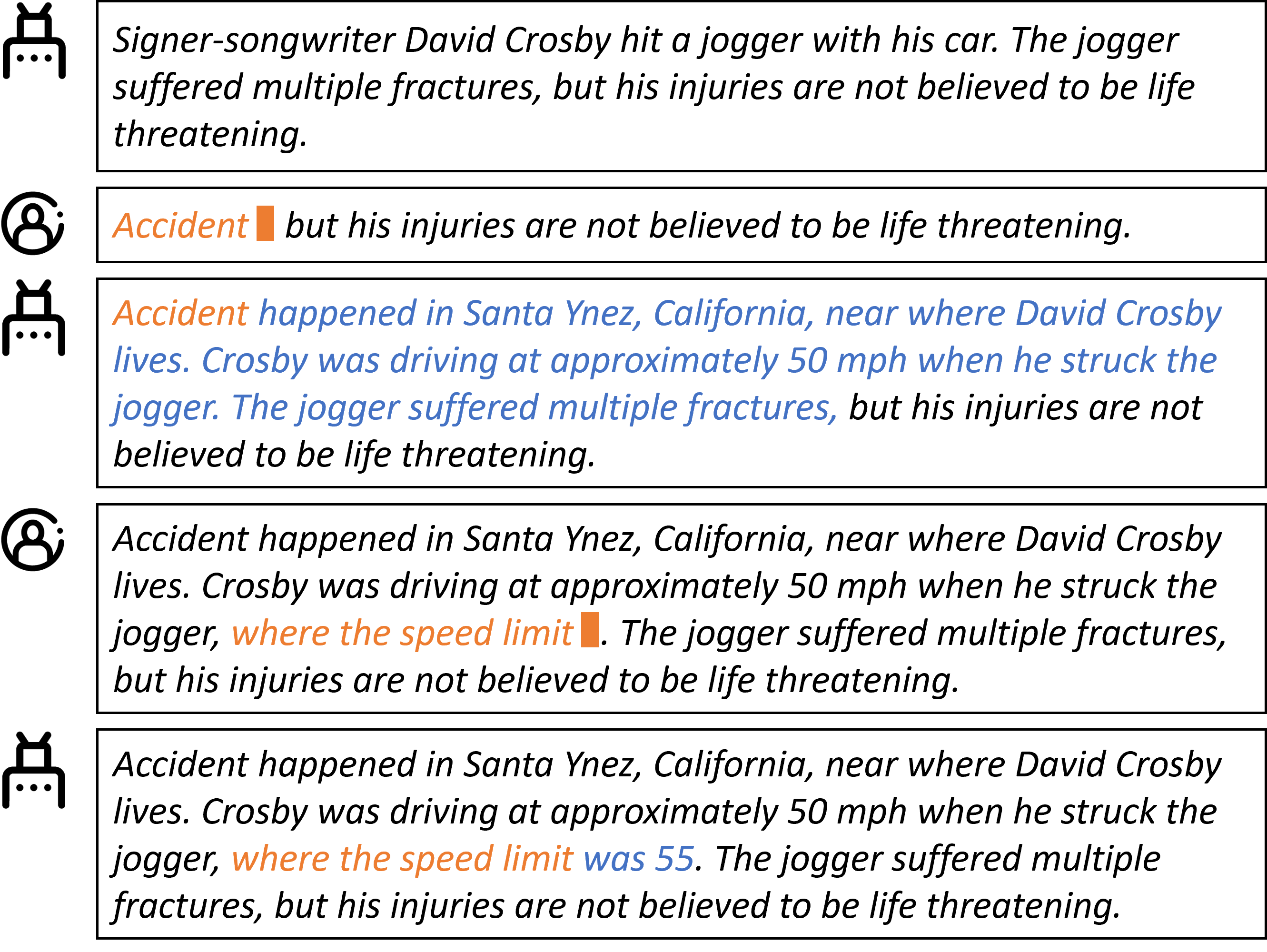}
  \caption{\label{fig:demo} Illustration on how \ModelName~interacts with human writer. }
\end{figure}

In this work, we present \ModelName -- Refinement and Editing Via Iterative Summarization Enhancement, a novel framework that transforms the summarization process into an interactive experience for human writers. Instead of generating static summaries, our approach enables users to efficiently edit and improve draft summaries iteratively, tailoring them to their specific needs and preferences. This results in the creation of high-quality, customized summaries that cater to individual requirements and contexts, moving beyond the limitations of traditional, one-size-fits-all summarization models. Figure \ref{fig:demo} shows an illustration. 



Our framework primarily consists of two models: one providing the initial draft summary and the other supporting the writer in refining the summary through interactive suggestions. We build upon a pretrained encoder-decoder summarization model and enhance its ability to generate contextually relevant and coherent suggestions for human edits. Through extensive experimentation and evaluation, we demonstrate the superior performance of our proposed framework in terms of salience, coherence, and adaptability to different situations.

The key innovation in \ModelName~lies in fostering a seamless collaboration between human writers and AI models, creating a truly interactive summarization experience. This interactive approach not only enables users to harness the power of AI to extract key information but also preserves the creativity and adaptability offered by human input. By empowering users to edit the summary iteratively until they are satisfied, our framework ensures the delivery of personalized, high-quality summaries tailored to diverse requirements.

We perform extensive human evaluations, suggesting the proposed framework can not only improve the efficiency of human editing, but also significantly enhance the summary quality.

\section{Related Works}

\noindent\textbf{Interactive summarization}.
There are several works exploring the how to facilitate human summary writing in an iterative way. For example, \citet{yan2011summarize} generate new summaries after their users to click on sentences they want to know more about. 
In \citet{avinesh2017joint} and \citet{avinesh2018sherlock}, users can indicate which bigrams of a candidate summary are relevant to their interests. 
The APRIL system \citep{gao2020preference} first let users to indicate preference between
candidate summaries, and then train a summary-ranking model to select the next pair of candidate summaries.
\citet{bohn2021hone} optimize document summaries for personal interest by collecting user feedback in the normal flow of reading, such as dwell time or gaze location.
More recently, \citet{shapira2021extending, shapira2022interactive} allow users to interactively submit queries in order to expand on the information on a topic in the summary.
In contrast, our framework is more versatile -- users can either specify their intent by the prompts, or let the model provide a few alternatives. 

\noindent\textbf{Text generation -- interactive editing}.
The task of human editing text interactively with AI is widely explored in other text generation tasks \citep{cheng2022mapping, lee2022evaluating}, e.g., machine translation \citep{barrachina2009statistical}. 
Many works are \textit{prefix-based}, i.e., new completions can only be generated left-to-right \citep{gonzalez2013interactive, peris2018active, peris2019interactive, peris2019online}. 
Few works \citep{gonzalez2016beyond, weng2019correct} allow edits at arbitrary positions. However, unlike \ModelName, the edits can only be words or sentences.







\noindent\textbf{Text infilling}.
There are two approaches for imbuing models with infilling capabilities: first, through new
architectures like SpanBERT \citep{joshi2020spanbert} and XLNet \citep{yang2019xlnet}. To list a few examples, XLNet modifies the attention mask in a standard transformer to enable token generation in any user-specified order, while Insertion Transformer \citep{stern2019insertion}, KERMIT \citep{chan2019kermit}, and
InDIGO \citep{gu2019insertion} allow the model to predict a location for the next token before predicting the token. Similarly, Blank Language models \citep{shen2020blank} generate text by iteratively selecting a blank and replacing it with a token (and optionally more blanks).

Alternatively, \citet{zhu2019text}, \citet{donahue2020enabling}, GLM \citep{du2022glm}, CM3
\citep{aghajanyan2022cm3}, InCoder \citep{fried2022incoder}, and \citet{bavarian2022efficient} utilize left-to-right autoregressive modeling by moving the infill regions to the end of context, with regions separated by sentinels. Notably, \citet{bavarian2022efficient} show the computational efficiency and superior performance of training in this way at scale. Our work extends \citet{bavarian2022efficient} to encoder-decoder models, and demonstrates the feasibility of its usage in summarization. 

Text infilling can also be performed using a GAN-based method \citep{fedus2018maskgan}, where REINFORCE is needed, or through gradient search \citep{liu2019tigs}.

\section{Method}

\begin{figure*}[tbh!]
  \centering  \includegraphics[width=0.9\linewidth]{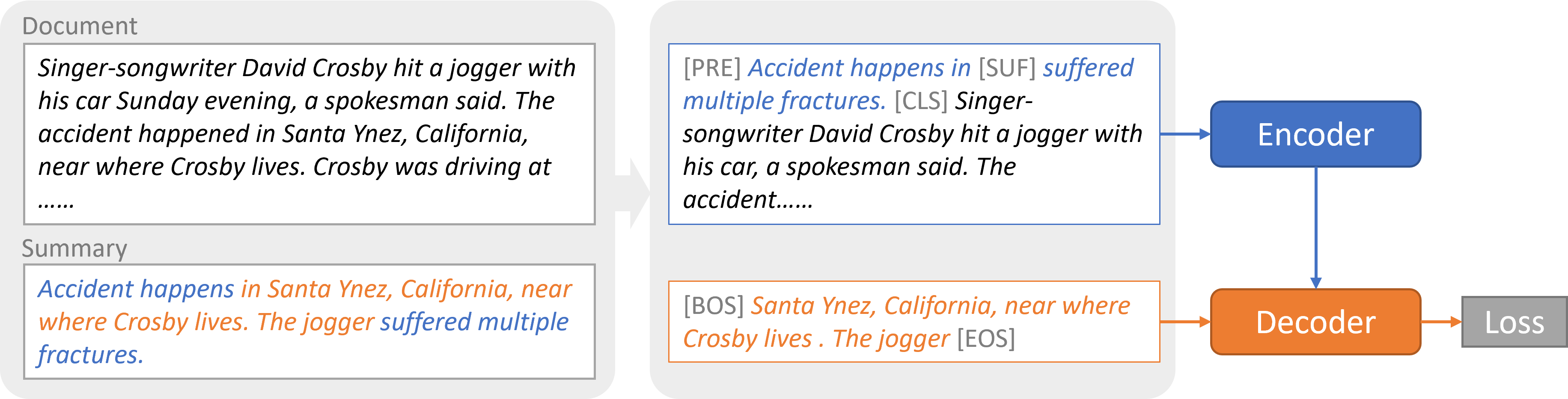}
  \caption{\label{fig:pipeline} Illustration on how the FIM model is trained. }
\end{figure*}

The key component of our framework is a Fill-In-the-Middle (FIM) model, which can provide alternatives to the human specified unsatisfactory part. Specifically, the input of the model is a source document sequence, a prefix sequence containing the summary before the deleted part and an optionally human start, a suffix sequence containing the summary after the deleted part. The goal of the model is to fill in a sequence that not only contains the key information of document, but also connect coherently with the prefix and the suffix. Following the advanced neural models for abstractive summarization  \citep{zhang2020pegasus, he2022z}, we adopt an encoder-decoder model architecture. 

\subsection{Training for FIM}

We start from a standard summarization training data $\mathcal{D}=\{d_i, t_i\}_{i=1}^N$, where $d_i$ is the $i$-th  document, $t_i$ is the corresponding summary, and they are both sequences of tokens. During training, we randomly divide the summary $t_i$ into three parts: the prefix $p_i$, the middle $m_i$, and the suffix $s_i$. Then, we concatenate the prefix, the suffix, and the document, together with their sentinel tokens, as the input for the encoder,
\begin{equation*}
    \texttt{[PRE]} \circ p_i \circ \texttt{[SUF]} \circ s_i \circ \texttt{[CLS]} \circ d_i,
\end{equation*}
where $\circ$ is the concatenation of tokens. We input the middle together with its sentinel tokens into the decoder,
\begin{equation*}
    \texttt{[BOS]} \circ m_i \circ \texttt{[EOS]}.
\end{equation*}
Here, we follow \citet{bavarian2022efficient} and adopt an \texttt{[EOS]} token to signal a successful concatenation of the middle and the suffix. The model is then trained with a standard cross-entropy loss for sequence-to-sequence models.

\subsection{Training for the Corner Cases}
\label{sec:train_corner}

Human edits can appear anywhere -- not only the middle, but also in the beginning and the end. In our preliminary experiments, we find that if the model is only trained for the data in the middle, it cannot handle the edits in the beginning and the end of the summaries. Especially, if the edits are in the end, the generation usually cannot reach \texttt{[EOS]} token, i.e., the generation cannot end. 

Therefore, we sample a proportion $\gamma$ of data for the edits in the beginning and the end. Specifically, we randomly split the summary $t_i$ into the prefix $p_i$ and the suffix $s_i$. On the one hand, to train the generation in the end, we input the concatenation of the prefix and the document into the encoder,
\begin{equation*}
    \texttt{[PRE]} \circ p_i \circ \texttt{[CLS]} \circ d_i,
\end{equation*}
and the suffix for the decoder,
\begin{equation*}
    \texttt{[BOS]} \circ s_i \circ \texttt{[EOS]}.
\end{equation*}
On the other hand, we exchange the position of $p_i$ and $s_i$ to train the generation in the beginning.


\section{Experiments}

\begin{table*}[htb!]
\centering
	\begin{adjustbox}{max width=0.8\textwidth}
\begin{tabular}{lllllll}
\toprule
Model                   & R-FIM       & R-begin     & R-end & $\ell_1$ & $\ell_2$ & R-all       \\ \hline
Proposed                & 51.79/30.30 & 49.51/28.87 & 37.36/20.31 &  -3.14  &  -2.97   &  43.32/20.96  \\
Context in decoder      & 45.72/25.62 & 44.25/24.72 & 36.32/19.41 &  -3.14 &  -3.84 & 43.37/20.94 \\
No corner case training & 51.86/30.24 & 46.67/27.11 & 26.97/12.67 &  -4.01  &  -3.04 & 10.67/2.42  \\
Base model              & 47.79/27.20 & 46.01/25.98 & 34.46/18.13 &  -4.32  &  -3.19  &  41.92/19.36 \\ \bottomrule
\end{tabular}
\end{adjustbox}
	\caption{ \label{tab:ablation}
		Different variants of the FIM model. Here, for R-FIM, R-begin, R-end, and R-all, the first number is ROUGE-1, the second number is ROUGE-2.
	}
\end{table*}

\subsection{Settings}

We use the CNN / Daily Mail (CNNDM) dataset \citep{see-etal-2017-get} for training and evaluation. We adopt the pretrained Z-Code++ \citep{he2022z} as the backbone encoder-decoder model. The proportion $\gamma$ of the training for the corner cases is $0.5$. We train the model for $10$ epochs with learning rate $7\times10^{-6}$. 

\subsection{Evaluation Metrics}

We evaluate the FIM model for three aspects:
\begin{enumerate}
    \item Is the generation salient in the document?
    \item Does the generation connect coherently with the rest of the summary?
    \item Can the model handle any possible positions?
\end{enumerate}
We propose to use three evaluation metrics for these three aspects, respectively. 

\noindent\textbf{ROUGE score}. We split the test set summaries of CNNDM into prefixes, middles, and suffixes. We feed the golden prefixes and  suffixes to the model, and compute the ROUGE scores \citep{lin2004rouge} between the generated texts and the golden middles. In this way, we measure whether the generated texts captures the important information as the golden summaries do.

\noindent\textbf{GPT Likelihood}. Large pretrained language models can be used to evaluate the coherence of text \citep{yuan2021bartscore}. Here, we adopt GPT-3.5 \citep{brown2020language}. Specifically, given a sequence of tokens $x=(x^{(1)}, \cdots, x^{(j)}, x^{(j+1)}, \cdots, x^{(N)})$, we evaluate whether $(x^{(1)}, \cdots, x^{(j)})$ are locally coherently connect with $(x^{(j+1)}, \cdots, x^{(N)})$ by computing the log-likelihood of generating $x^{(j+1)},x^{(j+2)}, \cdots$, given the previous tokens,
\begin{align*}
    & \ell_H ((x^{(1)}, \cdots, x^{(j)}), (x^{(j+1)}, \cdots, x^{(N)})) \\  = & \log p(x^{(j+1)}, \cdots, x^{(j+H)}|x^{(1)}, \cdots, x^{(j)}) \\
    = & \sum_{h=1}^H \log p_{\texttt{GPT}}(x^{(j+h)}|x^{(1)}, \cdots, x^{(j+h-1)})
\end{align*}
where $p_{\texttt{GPT}}(a^{(k)}|a^{(1)}, \cdots, a^{(k-1)})$ is the probability that next token is $a^{(k)}$ given previous token sequence $a^{(1)}, \cdots, a^{(k-1)}$, and the probability is provided by GPT model. Here, we adopt the likelihood on a fixed-length sequence $x^{(j+1)}, \cdots, x^{(j+H)}$ instead of the complete sequence $x^{(j+1)}, \cdots, x^{(N)}$ to alleviate the length effect.

For summarization, we consider the connectivity of the prefix-middle and the middle-suffix,
\begin{align*}
    \ell_1 & = \ell_H (p_i, m_i), \\
    \ell_2 & = \ell_H (p_i \circ m_i, s_i),
\end{align*}
where $p_i, m_i, s_i$ is from the split test set summary in the above section.  

\noindent\textbf{ROUGE score for corner cases}. Following the training recipe in Section \ref{sec:train_corner}, we split the test set summaries into the prefix and the suffix, ask the model to generate the prefix or the suffix accordingly, and compute the corresponding ROUGE scores.

\begin{table}[htb!]
\centering
	\begin{adjustbox}{max width=0.48\textwidth}
\begin{tabular}{lllll}
\toprule
Model                   & R-FIM       & R-begin     & R-end &  R-all       \\ \hline
Base model              & 47.79/27.20 & 46.01/25.98 & 34.46/18.13 &   41.92/19.36 \\
- DA          & 46.52/25.97 & 44.99/25.19 & 34.04/17.52 &  35.44/15.53 \\
- DA - RTD   & 46.22/25.80 & 45.15/25.22 & 33.40/17.02 &  39.90/18.21 \\ \bottomrule
\end{tabular}
\end{adjustbox}
	\caption{ \label{tab:pretrain}
		Different pretraining setting. 
	}
\end{table}

\subsection{Empirical Results}

\noindent\textbf{Different design choices for text infilling}.
We include three different variants of the proposed model in Table \ref{tab:ablation}. Our proposed training method can achieve the best performance in terms of salience and coherence. In Table \ref{tab:pretrain} we also show how the design choices in the pretraining stage affect the performance.

\noindent\textbf{Comparing to standard summarization model}. The last column of Table \ref{tab:ablation} shows the results when we do not provide summary context for generation, i.e., we ask the model to generate a complete summary. In comparison, in \citet{he2022z}, a vanilla finetuned summarization model can achieve a ROUGE-2 of 22.2. Our corner case training can significantly improve the generation quality.

\subsection{Human Evaluation}

\begin{figure}[t]
  \centering  \includegraphics[width=\linewidth]{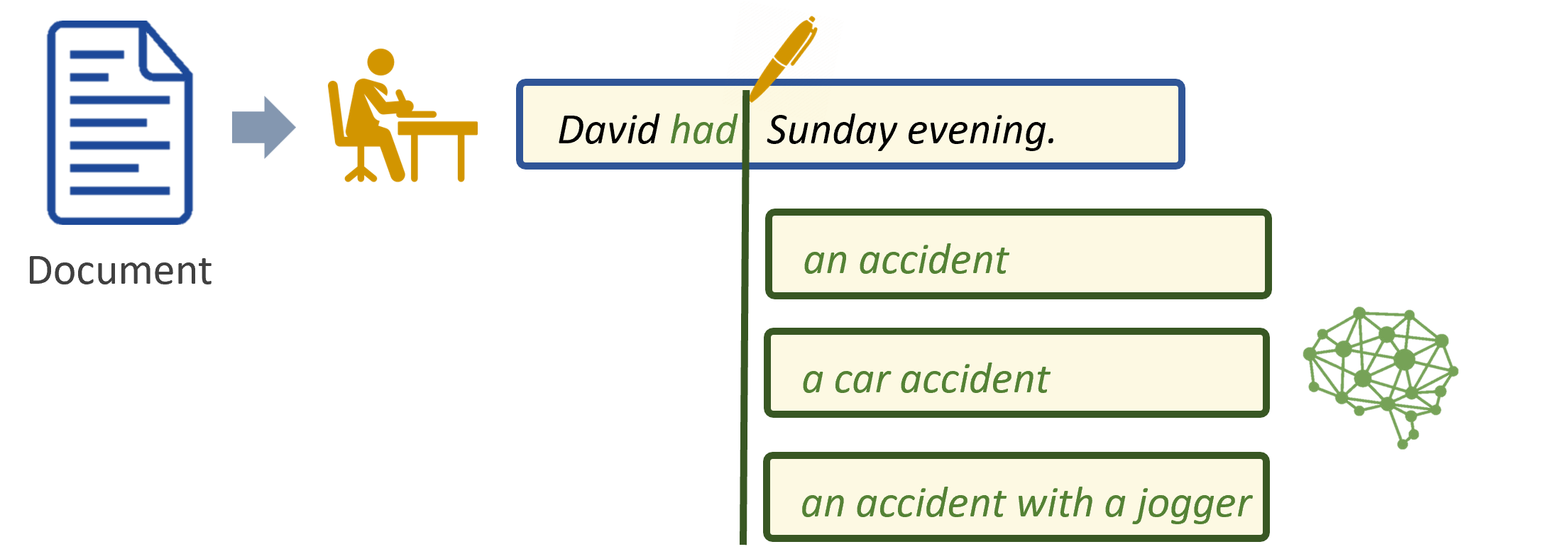}
  \caption{\label{fig:editing} Illustration on the human editing process with interaction. The annotators will be prompted with multiple suggestions of summary completions, and they choose one from them. }
\end{figure}

To validate our proposed pipeline can effectively help human writer to make edits, we conduct a contrast experiment on whether using REVIVE can improve editing efficiency. Specifically, we collect 120 document in three domains\footnote{We release the data together with the evaluation results at \url{https://github.com/microsoft/Interactive-Summarization}.} -- news, conversations, and blogs, and ask human annotators to edit the draft summaries until they are satisfied with the summary. The draft summaries are generated by the standard summarization model. We compare the editing processes with and without interaction. 

Figure \ref{fig:editing}  shows an illustration of the editing process with interaction. In practice, we prompt the annotators with 3 suggestions. We adopt the top-3 beams of beam search decoding method as suggestions, to enforce the suggestions are at least one-token different.

\begin{table*}[htb!]
\centering
	\begin{adjustbox}{max width=0.9\textwidth}
\begin{tabular}{lccccc}
\toprule
                  & Avg. Time  &  Avg. Rating & Accept / Accept w Edits / Reject & Hallucination Rate   \\ \hline
Draft Summary         &  --   & 3.99 $\pm$ 1.62 & 
    \begin{minipage}{.3\textwidth}
      \includegraphics[width=\linewidth]{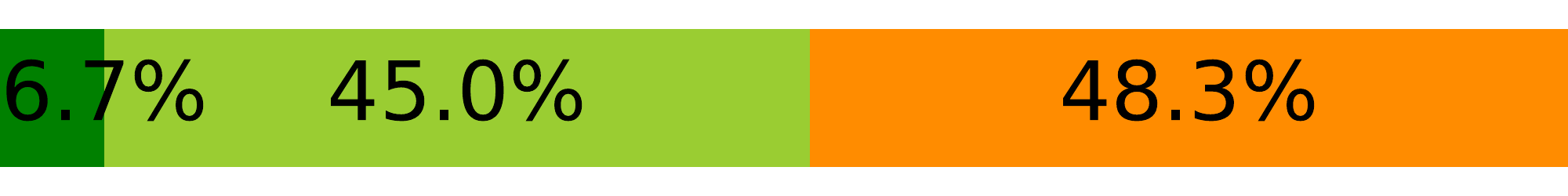}
    \end{minipage} & 0.12 \\
Human w/o interaction   & 903.0 & 4.61 $\pm$ 1.73 &
\begin{minipage}{.3\textwidth}
      \includegraphics[width=\linewidth]{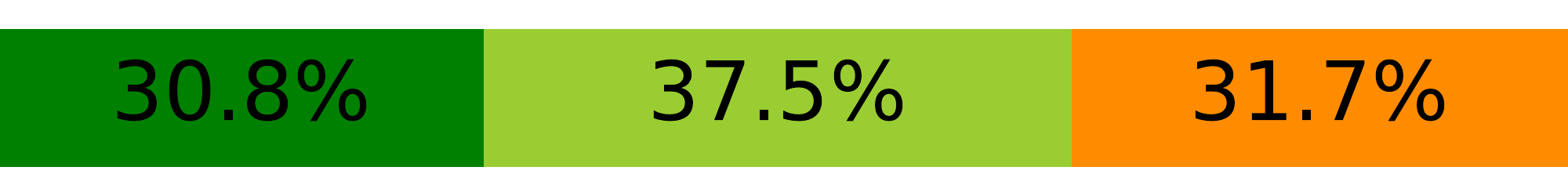}
    \end{minipage}& 0.04\\
Human w interaction     & 645.5 & 5.52 $\pm$ 0.84 &
\begin{minipage}{.3\textwidth}
      \includegraphics[width=\linewidth]{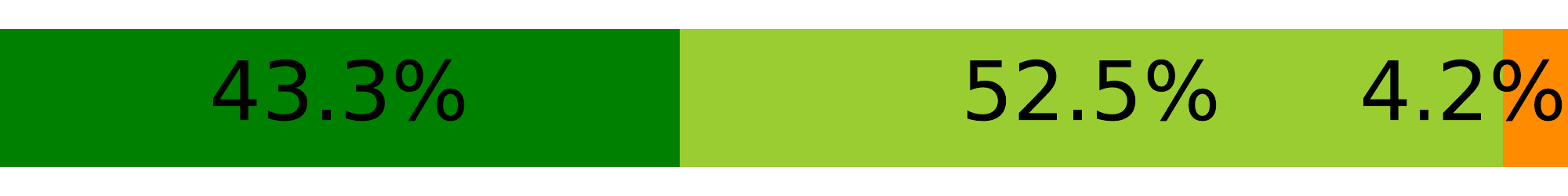}
    \end{minipage}& 0.03 \\
\bottomrule
\end{tabular}
\end{adjustbox}
	\caption{ \label{tab:human}
		Average editing time and annotation quality. The ratings are the general quality of the summary, with 1 meaning worst and 7 meaning best. 
	}
\end{table*}

The experiment consists of two stages. In the first stage, we perform a contrast experiment, to collect annotations with or without interaction. In the second stage, we evaluate the annotations collected from the first stage, to see whether the annotation quality and efficiency are improved. The experiment takes 3 annotators, where each annotator will 
\begin{enumerate}
    \item annotate 40 documents with interaction,
    \item annotate 40 documents without interaction, 
    \item evaluate the summaries of 40 documents, where each document has 3 summaries, i.e., the draft summary and the human annotated summary with and without interaction.
\end{enumerate}
The 40 documents in each of the above tasks are different for each annotator. In this way, we enforce the annotation time and evaluation results are unbiased, as the annotators will never see the same document across different tasks.

Table \ref{tab:human} shows REVISE can significantly save editing time and improve annotation quality. In practice, annotators trigger 2.2 suggestions for each document on average. Due to model inference latency and network latency, the suggestions will take 4.1s on average, which is also included in the annotation time reported in Table \ref{tab:human}. 

\section{Conclusion}

In this work, we propose \ModelName, offering an interactive and iterative approach for human writers to efficiently edit and refine summaries tailored to their specific needs. We adopt a fill-in-the-middle model for the encoder-decoder architecture underlies the framework, empowering it to generate coherent and contextually relevant suggestions for human edits, validated by a set of novel evaluation metrics. 
We perform extensive human evaluation, suggesting the proposed framework can improve both the editing efficiency and summary quality.
We hope \ModelName~can herald a new era in the summarization experience, with potentially transformative implications for practical applications and inspiring further research in interactive AI systems.


\bibliography{anthology,acl2023}
\bibliographystyle{acl_natbib}

\onecolumn
\appendix

\section{Human Annotation Guideline}
\label{sec:guideline}

Here we include the complete instructions for humen annotators, which include experiment details such as evaluation portal snapshots.

\subsection{Goal}
This experiment is to validate the effectiveness of the interactive summarization pipeline. Specifically, this pipeline is to facilitate the summary annotation process. It not only provides the annotators with a draft summary, but also interactively provides suggested completions to the annotators while they are editing the summary. This collected data will be used to further improve the annotation pipelines.

\subsection{Experiment Design and Instruction}
The experiment consists of two stages. In the first stage, we perform a contrast experiment, to collect annotations with or without interaction. In the second stage, we evaluate the annotations collected from the first stage, to see whether the annotation quality and efficiency are improved. The experiment needs 3 annotators, where each annotator will 
\begin{enumerate}
    \item annotate 40 documents with interaction,
    \item annotate 40 documents without interaction, 
    \item evaluate the summaries of 40 documents, where each document has 3 summaries.
\end{enumerate}

\subsubsection{Stage 1: Contrast experiment with or without interaction}

The experiment consists of two groups, i.e., the experimental group (with interaction) and the control group (without interaction), and there are 120 examples in each group. Since there are 3 annotators working on the experiment, each annotator will label 80 examples in total, where 40 belong to the experimental group and 40 belong to the control group. For each example, the annotator is given a source document, a draft summary, and possibly some interactive suggestions, and we will record 
\begin{enumerate}
    \item the annotated summary,
    \item the annotation time. 
\end{enumerate}
Since the annotation time is also critical to this experiment, we would like the annotators not to be distracted when doing annotation.

\vspace{10pt}
\noindent\textbf{A.2.1.1 Experimental Group}

\begin{figure}[t]
  \centering  \includegraphics[width=\linewidth]{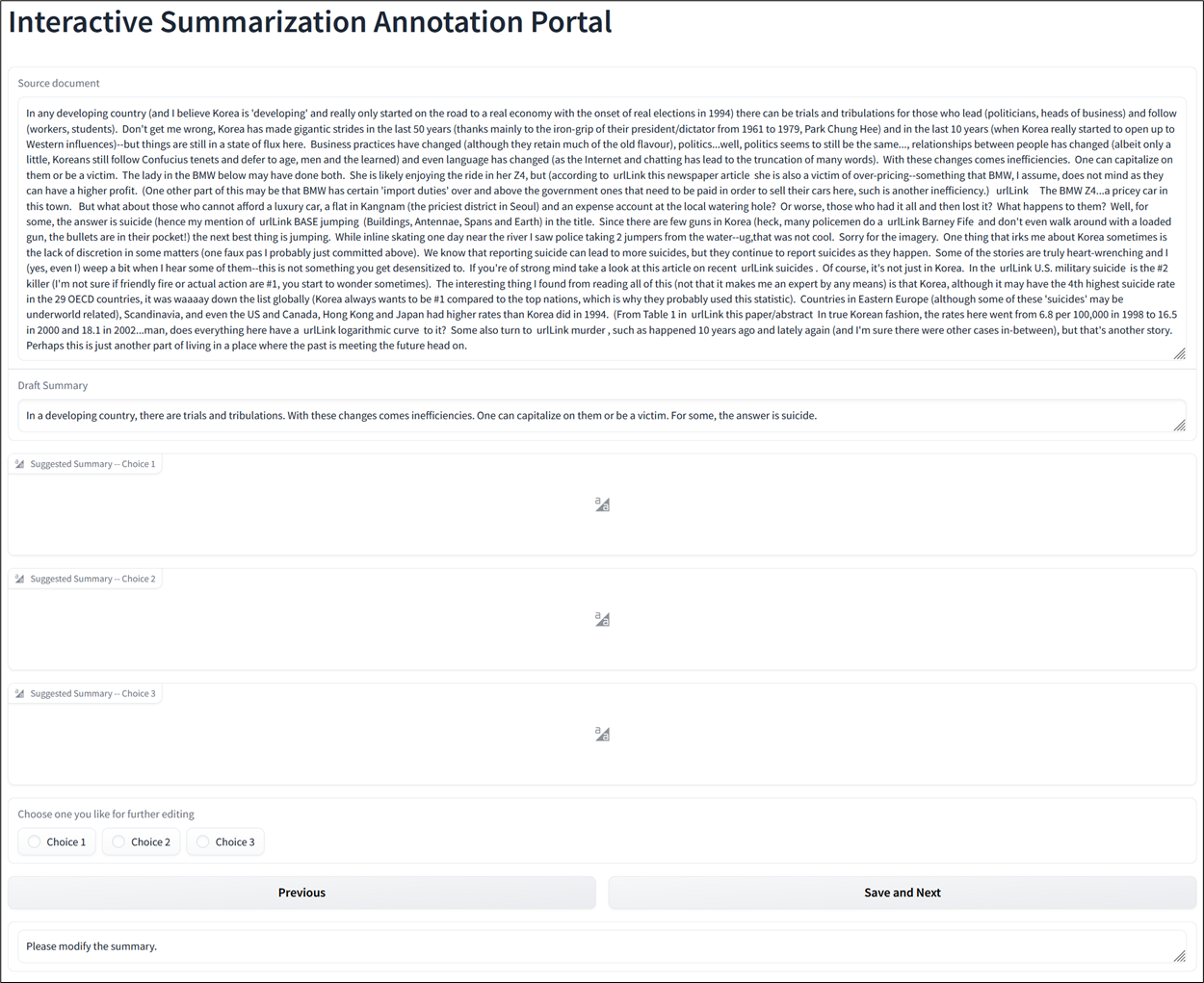}
  \caption{\label{fig:exp_portal} A snapshot of annotation portal for the experiment group. }
\end{figure}

\textbf{Portal description.} Figure \ref{fig:exp_portal} shows a snapshot of the annotation portal. 
The portal contains a few components:
\begin{enumerate}
    \item Source document: This is the document to be annotated.
    \item Draft Summary: To ease the annotation, we provide a draft summary as the start point of annotation. The draft summary may or may not be a good candidate summary. The annotator is asked to edit the draft summary until it satisfy the conditions described in the next section “Guidelines on What is a Good Summary Annotation”.
    \item Suggested summary – Choice 1/2/3: While the annotator is editing the summary, we provide three alternative suggestions interactively. Note that sometimes when the possible alternatives are limited, the suggestions can be the same.
    \item Choose one you like for further editing: If the annotator would like to adopt one of the suggestions, they can indicate so using this radio. After making a selection, the adopted suggestion will appear in the draft summary box for further editing.
    \item Save and Next button: Save current annotation and navigate to the next example. 
    \item Previous: Navigate to the previous example. The edits in the current example will be lost. If the annotator navigates back to a previous example, make edits and save the annotation again, the later annotation will be adopted.
\end{enumerate}

\textbf{Instruction on the usage of the suggestions.} The generation of the suggested summary is triggered by the ENTER key on the keyboard. The system will compare the old summary and the current summary and determine which part should be replaced.

As an illustrative example, if you are not satisfied with part of the summary, e.g., the highlighted part below,
\begin{figure}[h]
  \centering  \includegraphics[width=\linewidth]{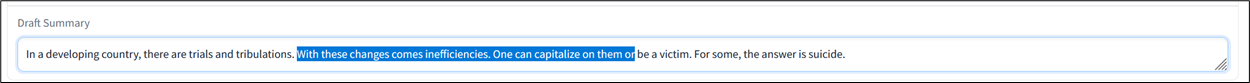}
\end{figure}
you can delete the highlighted part, hit the enter key, and it will trigger the system to generate suggestions, as shown below. 
\begin{figure}[h]
  \centering  \includegraphics[width=\linewidth]{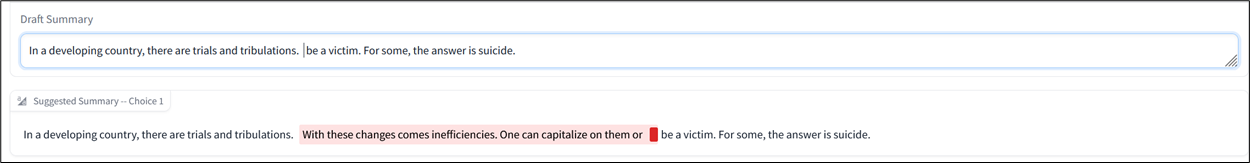}
\end{figure}
In this example, the generated suggestion is not good – it has the same content as the original one. To let the system know what you would like to insert instead, you can type the start you like, e.g.,
\begin{figure}[h]
  \centering  \includegraphics[width=\linewidth]{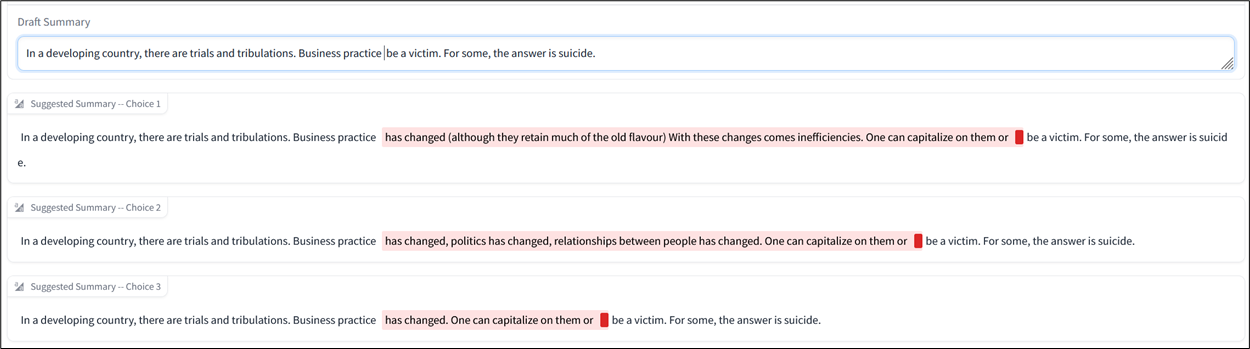}
\end{figure}
By typing “Business practice” into the summary box following up by two spaces, all suggestions will continue from “Business practice”. If you prefer the second choice, select “Choice 2”, and then the second suggestion will appear in the summary box for you to make further edits.
\begin{figure}[h]
  \centering  \includegraphics[width=\linewidth]{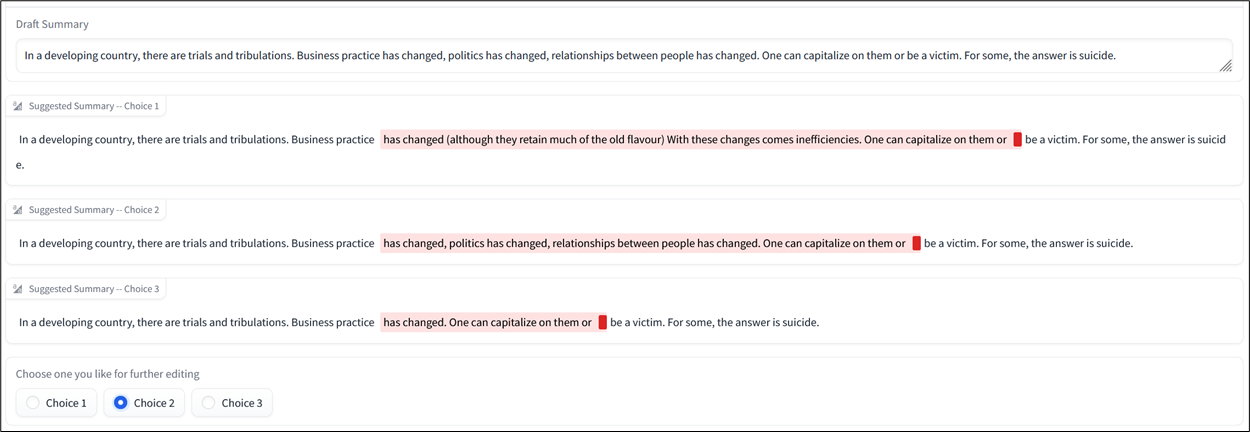}
\end{figure}

\vspace{10pt}
\noindent\textbf{A.2.1.2 Control Group}

\begin{figure}[t]
  \centering  \includegraphics[width=\linewidth]{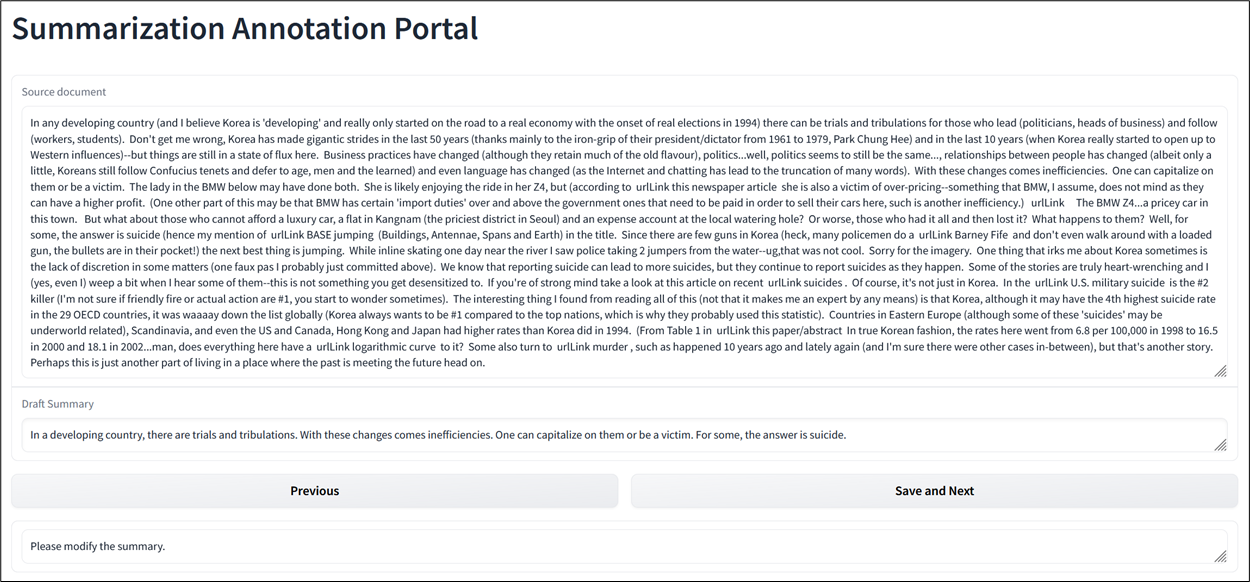}
  \caption{\label{fig:control_portal} A snapshot of annotation portal for the control group. }
\end{figure}

Figure \ref{fig:control_portal} shows a snapshot of the annotation portal. 

The purpose of source document box, draft summary box, previous button, and save \& next button is similar to the experimental group. The only difference is that in control group we do not provide interactive suggestions.

\vspace{10pt}
\noindent\textbf{A.2.1.3 More Guidelines on How to Make a Good Summary Annotation}

A good summary is a concise, fluent, and coherent text that can capture the main points of the original document/transcript accurately without any grammar errors, unrelated contents or hallucinations. Here are more detailed guidelines:
\begin{enumerate}
    \item Although a draft summary is provided, we would suggest reading through the document first to make a summary without being biased by the draft summary. The draft summary may or may not be trustworthy – it may miss the point completely or include nonexistent information.
    \item Keep perspective. If the document is of first-person perspective, use first-person perspective in the summary, too. Same applies to the third-person perspective.
    \item Summaries should cover important information in a concise way. It should avoid redundancy or unimportant information and should not be too verbose.
    \item Summaries should be self-contained and understandable by another readers without looking at the document. It should be fluent and coherent itself, like a very short passage.
    \item Summaries should not be too generic. For example, “This document talks about the economics of the developing countries.” is not as informative as “In a developing country, there are trials and tribulations.”
    \item Summaries should not add or make up new information that is not supported by the document.
\end{enumerate}

\subsubsection{Stage 2: Evaluation on Collected Summaries}
Each annotator will be assigned to 40 documents, each with 3 summaries. They will assign each summary a score on a 7-point Likert scale (1 = worst, 7 = best) and answer seven binary questions about potential issues in the summary.

\end{document}